\documentclass{article}

     \usepackage[preprint]{neurips_2023}

\usepackage[utf8]{inputenc} %
\usepackage[T1]{fontenc}    %
\usepackage[draft]{hyperref}       %
\usepackage{url}            %
\usepackage{booktabs}       %
\usepackage{amsfonts}       %
\usepackage{nicefrac}       %
\usepackage{microtype}      %
\usepackage{xcolor}         %
\usepackage{graphicx}
\usepackage{multirow}
\usepackage{tabularx}
\usepackage{IEEEtrantools}
\usepackage{color}
\usepackage{colortbl}
\usepackage[export]{adjustbox}
\usepackage{pifont}
\usepackage{amsmath}
\usepackage{amssymb}
\usepackage{booktabs}
\usepackage{authblk}
\title{IFT:~Image Fusion Transformer for Ghost-free High Dynamic Range Imaging}

\author[$\dagger$]{Hailing Wang\thanks{$\dagger$ First Author}\thanks{$\ddagger$ Corresponding Author}}
\author[$\dagger$]{Wei Li}
\author{Yuanyuan Xi}
\author{Jie Hu}
\author{Hanting Chen}
\author{Longyu Li}
\author[$\ddagger$]{Yunhe Wang}
\affil{Huawei Noah’s Ark Lab}
\affil[$\dagger$]{wanghailing6@huawei.com, wei.lee@huawei.com}
\affil[$\ddagger$]{yunhe.wang@huawei.com}

\makeatletter
\def\thanks#1{\protected@xdef\@thanks{\@thanks
		\protect\footnotetext{#1}}}
\makeatother

\begin{document}

\maketitle

\begin{abstract}
Multi-frame high dynamic range (HDR) imaging aims to reconstruct ghost-free images with photo-realistic details from content-complementary but spatially misaligned low dynamic range (LDR) images. Existing HDR algorithms are prone to producing ghosting artifacts as their methods fail to capture long-range dependencies between LDR frames with large motion in dynamic scenes. To address this issue, we propose a novel image fusion transformer, referred to as IFT, which presents a fast global patch searching (FGPS) module followed by a self-cross fusion module (SCF) for ghost-free HDR imaging. The FGPS searches the patches from supporting frames that have the closest dependency to each patch of the reference frame for long-range dependency modeling, while the SCF conducts intra-frame and inter-frame feature fusion on the patches obtained by the FGPS with linear complexity to input resolution. By matching similar patches between frames, objects with large motion ranges in dynamic scenes can be aligned, which can effectively alleviate the generation of artifacts. In addition, the proposed FGPS and SCF can be integrated into various deep HDR methods as efficient plug-in modules. Extensive experiments on multiple benchmarks show that our method achieves state-of-the-art performance both quantitatively and qualitatively.
\end{abstract}

\section{Introduction}
The dynamic range of typical real-world scenes varies across orders of magnitude (10$^6$:1, 120+dB)~\cite{radonjic2011dynamic}, which is far beyond the dynamic range that most consumer-grade image sensors can capture in a single shot~\cite{debevec2008recovering}. Consequently, over-exposed and~(or) under-exposed regions occur in the captured images.
A feasible solution is to capture several low dynamic range (LDR) images of the scene sequentially with different exposures and then blend valuable information from the LDR images into a target one. 
Nevertheless, due to object movement or camera shake, simply fusing multiple LDRs into an HDR one incurs unacceptable ghosting artifacts.
Even worse, when such occlusion occurs in saturated regions of the reference image, the fused HDR image is inevitably prone to severe ghosting artifacts.  

In order to address the aforementioned issue, a variety of traditional and deep methods have been proposed. Pece \emph{et al.} \cite{pece2010bitmap} designed binary operations to detect clusters of moving pixels within a bracketed exposure sequence, allowing for movements without generating ghosts. Li \emph{et al.} \cite{li2014selectively} proposed to remove ghosts in the LDR domain and introduced a content adaptive bilateral filter for detail extraction and fusion. Though these shallow methods mitigate the generation of artifacts, their HDR imaging performance is limited by the inability to accurately localize the range of motion due to large-scale object movement in the scene. More recently, with the great performance improvement of deep learning methods, researchers began to experiment with deep HDR methods \cite{ram2017deepfuse, kalantari2017deep, prabhakar2019fast, wu2018deep, yan2019attention, yan2019multi, yan2020deep, dai2021waveletbased, HDR_GAN_2021, ADNet}. According to the alignment of the moving region of LDRs, these methods can be simply divided into two categories. One is to explicitly align the LDR images involving moving objects by adopting the optical flow technique, such as \cite{kalantari2017deep, prabhakar2019fast}, assuming that the brightness of a moving object in different LDR images does not change. However, the above assumption does not hold for HDR tasks due to the varied exposure of each frame of LDRs, which easily leads to ghosting in the result. Another way is to implicitly align moving objects in LDR frames with the aid of Unet design \cite{yan2019multi} and spatial attention \cite{yan2019attention}, circumventing the exposure consistency restriction of the optical flow technique. Nevertheless, these methods typically have small receptive fields, which disable global modeling to a certain extent.

Recently, Transformer has gained a surge of interest for image processing. Chen \textit{et al.} proposed the first pre-trained image processing transformer IPT~\cite{chen2021pre}. SwinIR~\cite{liang2021swinir} first applies the residual Swin Transformer block to the image restoration for deep feature extraction. 
Thanks to its inherent superior long-range modeling capability, the visual transformer  \cite{chen2021pre, wang2022uformer} has shown great potential in HDR imaging. %
Liu \emph{et al.} \cite{liu2022ghost} proposed the first transformer-based framework named HDR-Transformer (HDR-T), which achieves superior performance for HDR imaging than previous methods. However, HDR-T inherits the spatial attention paradigm from AHDR~\cite{yan2019attention} and only performs attention modeling on limited-sized windows and does not exert the long-range modeling ability of transformer. Subsequently, while Song \emph{et al.} \cite{song2022selective} proposed a transformer-based selective HDR image reconstruction network with the guidance of a ghost region detector, the ghost region detector is inaccurate and requires proper threshold selection, which means artifacts are easily generated in the results. 

\begin{figure}[t]
	\centering
	\includegraphics[width=0.9\textwidth, trim=0 10 0 0,clip]{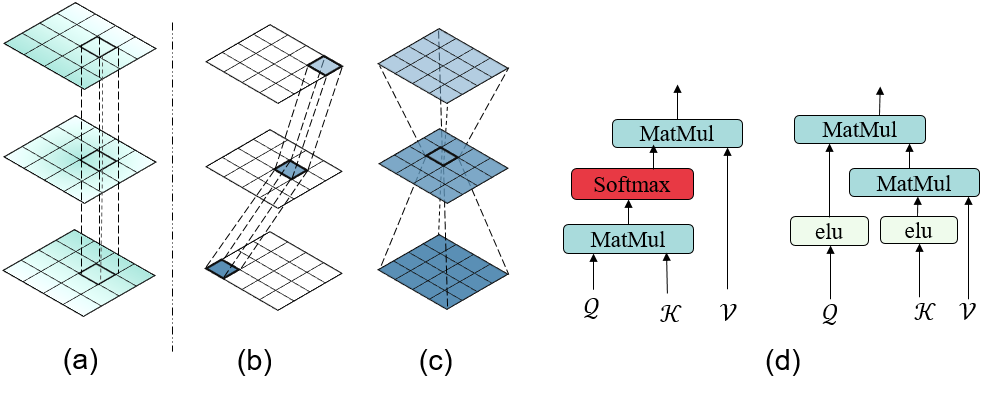}
	\caption{(a) Mainstream HDR fusion scheme (b)(c) Our fusion scheme (d) Dot-Product Attention
 and Linear Attention.}
	\label{sub}
	\vspace{-7mm}
\end{figure}

To fully exploit transformers‘ advantages in long-range dependency modeling, we propose a novel image fusion transformer (IFT), which draws long-range attention from supporting LDR images for dynamic scenes.
Specifically, the proposed IFT consists of two main components, including a fast global patch searching module (FGPS) and a self-cross patch fusion transformer (SCF). The FGPS module performs matching of similar patches between frames, as shown in Figure \ref{sub} (b), to match the large-scale motion of the same object between different frames. This module can search for similar blocks globally and establish the correspondence of similar blocks between frames. In addition, the SCF module performs feature fusion on the matched similar blocks in a linear computational way. As shown in Figure \ref{sub} (c), on the inter-frame similar patches, the reference frame performs global attention calculation with the support frames. Compared with the existing mainstream HDR fusion scheme, that is, to consider whether the corresponding position of the relationship between frames is a semantically consistent area at the pixel level, then to fuse information under a limited receptive field as shown in Figure \ref{sub} (a), the main contributions of this work can be summarized as follows:

\begin{itemize}
	\item We propose a novel image fusion transformer, termed IFT, for multi-exposure HDR imaging, which contains a FGPS module for extracting long-range dependencies of moving objects and a SCF transformer for fusing intra-frame and inter-frame features. 
	
	\item The proposed FGPS and SCF can be integrated into various deep HDR methods as efficient plug-in modules to globally fuse multi-frame information with linear complexity to input resolution.
	
	\item Extensive experiments conducted on representative HDR benchmarks demonstrate that our IFT algorithm achieves state-of-the-art performance. 
	
\end{itemize}

\section{Related Work}
In this section, we briefly summarize existing multi-exposure HDR reconstruction methods and transformer-based methods for low-level vision tasks.

\subsection{HDR Fusion from Multi-exposures}
\textbf{Traditional HDR Fusion.} 
Many traditional methods \cite{debevec1997, khan2006ghost, grosch2006fast, jacobs2008automatic, pece2010bitmap, heo2010ghost, sen2012robust, hu2013hdr, li2014selectively, li2022sj} have been proposed for multi-exposure HDR imaging, which can be roughly divided into two categories, including rejection-based and alignment-based approaches. The former methods assume that images can be globally registered so that each pixel can be classified as static or ghosted. The ghost pixels can then be rejected, leaving only the static ones for further fusion to reconstruct the ghost-free HDR images. For example, Grosch \emph{et al.} \cite{grosch2006fast} designed an error map for moving objects with the aid of color difference of inputs to reject ghost pixels. Sidibe \emph{et al.} \cite{sidibe2009ghost} utilized the prior knowledge that the pixel value in static area usually increases with the increase of exposure to distinguish between the static and ghost areas. Heo \emph{et al.} \cite{heo2010ghost} employed a generalized weighted filtering technique to estimate reliable radiance values for each exposure, based on the global intensity transfer functions obtained from joint probability density functions between different exposure images. On the contrary, the alignment-based methods aim to align the non-reference LDR images to the reference one before fusion. For instance, Tomaszewska \emph{et al.} \cite{tomaszewska2007image} suggested using the SIFT method to search for key points in order to identify matrices for transforming images and eliminating global misalignments. In contrast to the key points searching strategy, some methods introduced optical flow to align LDR images, further enhancing the performance of HDR imaging. In addition, some techniques based on patch alignment have been proposed. For example, Sen \emph{et al.} \cite{sen2012robust} constructed a patch-based energy-minimization formulation and concurrently incorporated the LDR image alignment and HDR image merging into the optimization.  

\textbf{Deep HDR Methods.}  
With the great success of deep learning, many CNN-based approaches \cite{ram2017deepfuse, kalantari2017deep, prabhakar2019fast, wu2018deep, yan2019attention, yan2019multi, yan2020deep, dai2021waveletbased, HDR_GAN_2021, ADNet} have been proposed for HDR imaging. For example, Kalantari \emph{et al.}\cite{kalantari2017deep} introduced optical flow into the design of convolutional networks and proposed the first CNN-based method for HDR reconstruction. Taking the optical flow approach into the design of CNN, however, is prone to incur inaccurate alignment for LDR frames. The potential reason is that the optical flow assumes that the brightness of the corresponding pixels between different frames should be consistent, which is hard to satisfy in real-world applications. Subsequently, many deep methods \cite{yan2019attention, yan2019multi} have been proposed to implicitly align the LDR frames to reduce HDR artifacts. Yan \emph{et al.} \cite{yan2019attention}, for example, applied spatial attention mechanism for alignment of LDR frames. Due to the limited receptive field of convolutional networks, it is usually difficult for CNN-based models to accurately distinguish which pixel should be aligned since they fail to capture global semantic information. As a result, these methods typically need to elaborately design an appropriate fusion network in order to mitigate the misalignment issue during the subsequent fusion process.   

\subsection{Low-Level Vision Transformer}
Recently, vision transformer has achieved remarkable success in computer vision tasks, such as object detection \cite{carion2020end,zhu2020deformable,fang2021you}, image recognition \cite{dosovitskiy2020image, wang2021not}, semantic segmentation\cite{xie2021segformer}. Subsequently, numerous transformer-based modules have been proposed to resolve low-level visual tasks and produce better visual results because the multi-head self-attention mechanism in transformer models is proved to be excellent at modeling long-range dependency among image contexts. For instance, Chen \emph{et al.} \cite{chen2021pre} proposed a transformer architecture called IPT, which provides a pre-trained model for different low-level vision tasks. Followed by IPT, Wang \emph{et al.} \cite{wang2022uformer} developed a hierarchical encoder-decoder network based on transformer blocks, which reduces computational complexity while improving performance. During the same period, Zamir \emph{et al.} \cite{zamir2022restormer} designed a Restormer, which utilizes an encoder-decoder transformer and achieves state-of-the-art performance on several image restoration tasks. 
In the realm of HDR imaging, the first HDR transformer, a construct aimed at the reconstruction of premium-quality, ghost-free HDR images, was innovatively engineered by Liu \textit{et al.} ~\cite{liu2022ghost}, drawing inspiration from the research outlined in~\cite{liu2021swin}. Song \textit{et al.}~\cite{song2022selective} further contributed to the field by proposing a transformer-influenced selective HDR image reconstruction network, directed by the implementation of a ghost region detector. It must be noted, though, that these methodologies often exhibit limited receptive fields in their respective attention modules, thereby partially limiting their global modeling capacities.
\begin{figure*}[t]
	\centering
	\includegraphics[width=0.95\textwidth]{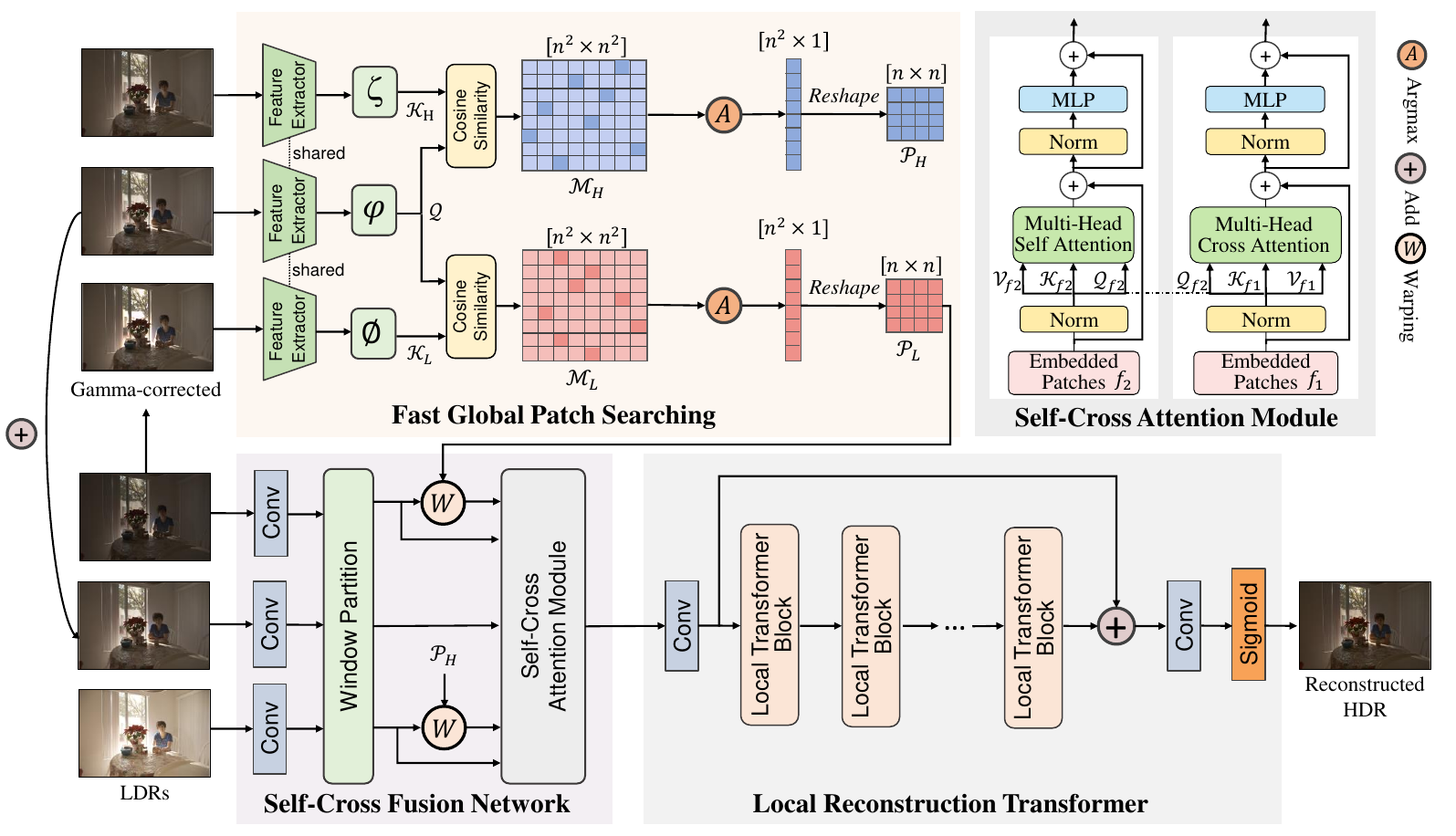}
	\caption{The architecture of the proposed IFT. It consists of three sub-network, semantic-aware patch searching, self-cross fusion network and local reconstruction transformer.}
	\label{fig_main}
	\vspace{-4mm}
\end{figure*}
\section{Methodology}
Given a reference LDR image $I_2$ (medium exposure) and two supporting LDR images $I_1$ (short exposure) and $I_3$ (long exposure), we attempt to design an end-to-end model to generate a ghost-free HDR image. 
An overview of our proposed architecture is presented in Fig.\ref{fig_main}, which contains a fast global patch searching module (FGPS) for semantic alignment of moving objects, a self-cross fusion network (SCF) for similar semantics fusion, and a local reconstruction transformer (LRT) for context detail recovery. 
\subsection{Fast global patch searching module.}
In order to extract the long-range dependencies for compensating for large motions of objects, the FGPS module seeks out the patch from supporting frames that has the closest semantics to each patch of the reference frame. Specifically, the input LDR frames are first preprocessed by gamma correction noted as $I_1^\gamma$, $I_2^\gamma$, $I_3^\gamma$ in order to avoid the influence of illumination. 
Then, the corrected LDR frames are input to parameter-shared feature extractors $\mathcal{FE}$ for downsampling, generating three corresponding feature maps $f_1^\gamma$, $f_2^\gamma$, $f_3^\gamma$:
\begin{equation}
	f_1^\gamma=\mathcal{FE}\left(I_1^\gamma\right),
	f_2^\gamma=\mathcal{FE}\left(I_2^\gamma\right),
	f_3^\gamma=\mathcal{FE}\left(I_3^\gamma\right)
\end{equation}
Note that, the applied parameter-shared feature extractors map three LDR frames into the same mapping space, so as to prevent semantic misalignment during the patch searching and matching stage. Subsequently, we use three embedding networks $\phi(\cdot)$, $\varphi(\cdot)$
and $\zeta(\cdot)$ to extract tokens. The queries $\mathcal{Q}$ and keys $\mathcal{K}_L$ and $\mathcal{K}_H$ are extracted by $\phi(\cdot)$, $\varphi(\cdot)$, $\zeta(\cdot)$ and denoted as
\begin{equation}
	\begin{aligned}
		\mathcal{Q}=\phi\left(f_2^\gamma\right)=\left\{q^i, i \in[1, n]\right\},
		\\
		\mathcal{K}_L=\varphi\left(f_1^\gamma\right)=\left\{{k_l}^i, i \in[1, n]\right\},
		\\
		\mathcal{K}_H=\zeta\left(f_3^\gamma\right)=\left\{{k_h}^i, i \in[1, n]\right\},
	\end{aligned}
\end{equation}
respectively, where $n$ represents the number of tokens.
We then calculate the cosine similarity between the queries and keys, resulting in two attention maps, denoted as $\mathcal{M}_{L}$ and $\mathcal{M}_{H}$ between the reference and supporting patches that help us identify the long-range dependencies of moving objects across LDR frames, which can be expressed as:
\begin{equation}
	\mathcal{M}_{L}=\underset{1 \ldots n}{\operatorname{argmax}}\left\langle\frac{\mathcal{Q}^n}{\left\|\mathcal{Q}^n\right\|_2^2}, \frac{\mathcal{K}_L^n}{\left\|\mathcal{K}_L^n\right\|_2^2}\right\rangle
\end{equation}

\begin{equation}
	\mathcal{M}_{H}=\underset{1 \ldots n}{\operatorname{argmax}}\left\langle\frac{\mathcal{Q}^n}{\left\|\mathcal{Q}^n\right\|_2^2}, \frac{\mathcal{K}_H^n}{\left\|\mathcal{K}_H^n\right\|_2^2}\right\rangle
\end{equation} 
After that, we apply a argmax operation $\mathcal{A}$ on the attention maps to obtain two position maps $\mathcal{P}_{L}$, $\mathcal{P}_{H}$ that store location indexes with the closest semantics for supporting frames, which can be defined as:
\begin{equation}
	\mathcal{P}_{L}=\mathcal{A}(\mathcal{M}_{L}), \mathcal{P}_{H}=\mathcal{A}(\mathcal{M}_{H})
\end{equation} 
To be more precise, in the position maps, each element representing a supporting patch saves the index of the reference patch that shares the greatest semantic similarity to that supporting patch. Moreover, since softmax is not differentiable during optimization, we use the straight through estimator (STE) to skip the gradient backpropagation of argmax and directly update the gradient for feature extractors. 

It should be noted that in the downsampling process, the original LDR frames are first sampled down to the size of $256 \times 256$ and then extracted by feature extractors to generate small-sized feature maps for attention calculation. This operation of finding the semantically most similar blocks in the down-sampled semantic space greatly reduces the computational effort and speeds up training and inference. In fact, many transformer-based models tailored to advanced visual tasks such as image recognition and detection have applied feature downsampling to reduce computing consumption. There are, however, very few models that incorporate feature downsampling for low-level visual tasks like image denoising and super-resolution. The primary reason is that feature downsampling will remove image details, which imposes limited effects on semantic information for advanced visual tasks but poses a destructive risk to the loss of context details for low-level visual tasks. However, in our model, we can adopt the downsampling operation to reduce computation complexity before patch searching and matching since the proposed FGPS network only focuses on the position of semantic  patches rather than the local pixel details.

\subsection{Self-Cross Fusion Transformer}
With the position maps obtained by FGPS, the SCF network is constructed to perform intra-frame and inter-frame feature fusion on the semantic-similar patches, so as to fully integrate the semantic-aligned information for further reconstruction of local details. Specifically, the SCF takes three original LDR frames $I_1, I_2, I_3$ and their corresponding gamma corrected versions $I_1^\gamma, I_2^\gamma, I_3^\gamma$ as inputs, and then sends them into three different convolution networks $\{\operatorname{conv1}, \operatorname{conv2}, \operatorname{conv3}\}$ to generate three feature maps $f_1, f_2, f_3$: 
\begin{equation}
	\begin{aligned}
		f_1=\operatorname{conv1}\left(I_1, I_1^\gamma\right),\\
		f_2=\operatorname{conv2}\left(I_2,I_2^\gamma\right),\\
		f_3=\operatorname{conv3}\left(I_3,I_3^\gamma\right), 
	\end{aligned}
\end{equation}

In order to correctly fuse the features of matched patches cross different frames, it is important and necessary to align the semantic patches of supporting frames to those of the reference frame, especially for the patches containing or situated on moving objects. To this end, we apply the obtained position maps $\mathcal{P}_{L}$  and $\mathcal{P}_{H}$ to alter the patch positions for the two supporting frames so as to align the semantics with the reference frame. This process can be achieved by a simple warping operation $\mathcal{W}$:
\begin{equation}
	f_1^w = \mathcal{W}(f_1, \mathcal{P}_{L}),\quad f_3^w = \mathcal{W}(f_3, \mathcal{P}_{H})
\end{equation} 
where $f_1^w$ and $f_3^w$ represent the warped feature maps of $f_1$ and $f_3$, correspondingly. To keep enough context details, the original feature maps $f_1, f_2, f_3$ and the warped feature maps $f_1^w, f_3^w$ are all sent to a self-cross attention transformer ($\operatorname{SAAT}$) for final feature fusion, as follows:
\begin{equation}
	\hat{f}_1, \hat{f}_1^w, \hat{f_2}, \hat{f_3}, \hat{f}_3^w = \operatorname{SAAT}(f_1, f_1^w, f_2, f_3, f_3^w)
\end{equation}

In the $\operatorname{SAAT}$, we use embedding network $\theta_0(\cdot)$ to extract queries $\mathcal{Q}_{f_2} = \theta_0(f_2)$, $\theta_f(\cdot)$ to generate keys $\mathcal{K}_f = \theta_f(f)$, and $\vartheta_f(\cdot)$ to obtain values $\mathcal{V}_f = \vartheta_f(f)$  for $\forall f \in \{f_1, f_1^w, f_2, f_3, f_3^w\}$. When the keys $\mathcal{K}_f$ come from the reference frame $I_2$, the $\operatorname{SAAT}$ actually calculates a self-attention as shown in the left side of Figure \ref{sub} (d):
\begin{equation}
	\hat{f_2} = \operatorname{SoftMax}\left(\mathcal{Q}_{f_2} \mathcal{K}_{f_2}^T / \sqrt{d}\right) \mathcal{V}_{f_2}
\end{equation}
When the keys $\mathcal{K}_f$ come from the supporting frames $\{I_1, I_3\}$, the $\operatorname{SAAT}$ calculates a cross-attention:
\begin{equation}
	\hat{f}=\operatorname{SoftMax}\left(\mathcal{Q}_{f_2} \mathcal{K}_{f}^T / \sqrt{d}\right) \mathcal{V}_{f}, 
	\forall f \in \{f_1, f_1^w, f_2, f_3, f_3^w\}
\end{equation}
However, the computation complexity of self-attention and cross-attention are O($N^2$). Linear Transformer \cite{katharopoulos2020transformers} proposes to reduce the computational complexity to O(N) by substituting the  kernel function $sim(Q, K)=\epsilon(\mathcal{Q}) \epsilon(\mathcal{K})^T$ for the indicator kernel $\operatorname{SoftMax}$ used in the original attention layer, where $\epsilon() = elu() + 1$. This operation is illustrated by the computational graph in the right side of Figure \ref{sub} (d). The multiplication between $\epsilon(K)^T$ and $\mathcal{V}$ can be carried out first and the computation cost is reduced to O(N). Thus, we redefine self and cross attention module:
\begin{equation}
	\hat{f}=(elu(\mathcal{Q}_{f_2}) + 1)\left( (elu(\mathcal{K}_{f}) + 1)\right) \mathcal{V}_{f}^T, 
	\forall f \in \{f_1, f_1^w, f_2, f_3, f_3^w\}
\end{equation}
 After that, the five fused features $\{\hat{f}_1, \hat{f}_1^w, \hat{f_2}, \hat{f_3}, \hat{f}_3^w\}$ are concated together for further reconstruction.

\subsection{Local Reconstruction Transformer \& Loss Function}
\textbf{Local Reconstruction Transformer} Based on the fused features obtained by the SCF, we aim to generate a ghost-free HDR image using a local reconstruction transformer (LRT), which adopts the CA-ViT \cite{liu2022ghost} as a basic block. The CA-ViT block contains two branches, including a transformer branch and a channel attention branch. The former is constructed by a multi-head self-attention (MSA) and a multi-layer perceptron (MLP) with residual connection, while the latter contains two linear layers followed by a ReLU and a sigmoid activation layer. These two branches are finally combined by a element-wise addition to reconstruct ghost-free HDR images.

\textbf{Loss Function} To guide the training of IFT for producing desired tonemapped results, the loss function takes into account the tonemapped domain using the commonly used $\mu-law$ function:
\begin{equation}
	\mathcal{T}(\mathcal{O})=\frac{\log (1+\mu \mathcal{O})}{\log (1+\mu)}
\end{equation}
where $\mathcal{T}(\mathcal{O})$ is the tonemapped HDR image, and we set $\mu$ to 5000. 

Consistent with the previous method \cite{liu2022ghost}, we also utilizes $\mathcal{L}_1$ loss and perceptual loss $\mathcal{L}_p$ to optimize the proposed IFT. As a result, the loss
function can be written in the following:
\begin{equation}
	\mathcal{L}_{all}=\mathcal{L}_1+\lambda \mathcal{L}_p
\end{equation}
We use $\mathcal{L}_1$ loss to measure the distance between HDR image result $\mathcal{O}$ and the ground truth HDR image $\mathcal{O}_{gt}$, which can be defined as
\begin{equation}
	\mathcal{L}_r=\left\|\mathcal{T}\left(\mathcal{O}\right)-\mathcal{T}\left(\mathcal{O}_{gt}\right)\right\|_1
\end{equation}
In addition, we adopt the perceptual loss\cite{johnson2016perceptual} to measure the high-level feature similarity and achieve satisfied performance. The perceptual loss can be described as the following formula:
\begin{equation}
	\mathcal{L}_p=\sum_{j \in \mathcal{J}}\left\|\Phi_j\left(\mathcal{T}\left(\mathcal{O}\right)\right)-\Phi_j\left(\mathcal{T}\left(\mathcal{O}_{gt}\right)\right)\right\|_1
\end{equation}
where $\Phi_j(z)$ is the j-th layer feature map of $z$ from pre-trained high-level semantic feature extractor network $\Phi$, and $\mathcal{J}$ denotes a set of chosen layers. In this paper, we adopt VGG-16\cite{simonyan2014very} that trained on ImageNet for image classification as the feature extractor network.

\begin{figure*}[t]
	\centering
	\includegraphics[width=1\textwidth, trim=10 224 54 6,clip]{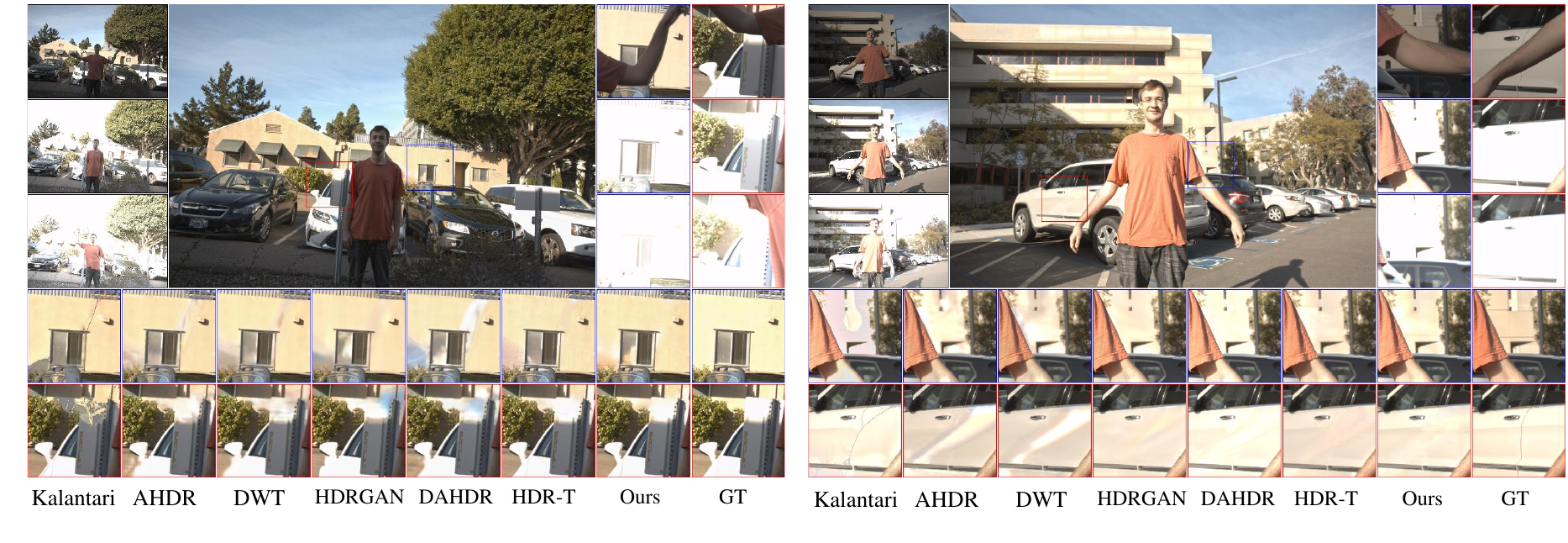}
	\vspace{-6mm}
	\caption{Comparison of qualitative results of the proposed method with the state-of-the-art methods on Kalantari dataset.  Zoom in for better observation.}
	\label{main_vision}
	\vspace{-3mm}
\end{figure*}
\begin{figure*}[t]
	\centering
	\includegraphics[width=1\textwidth, trim=10 224 54 6,clip]{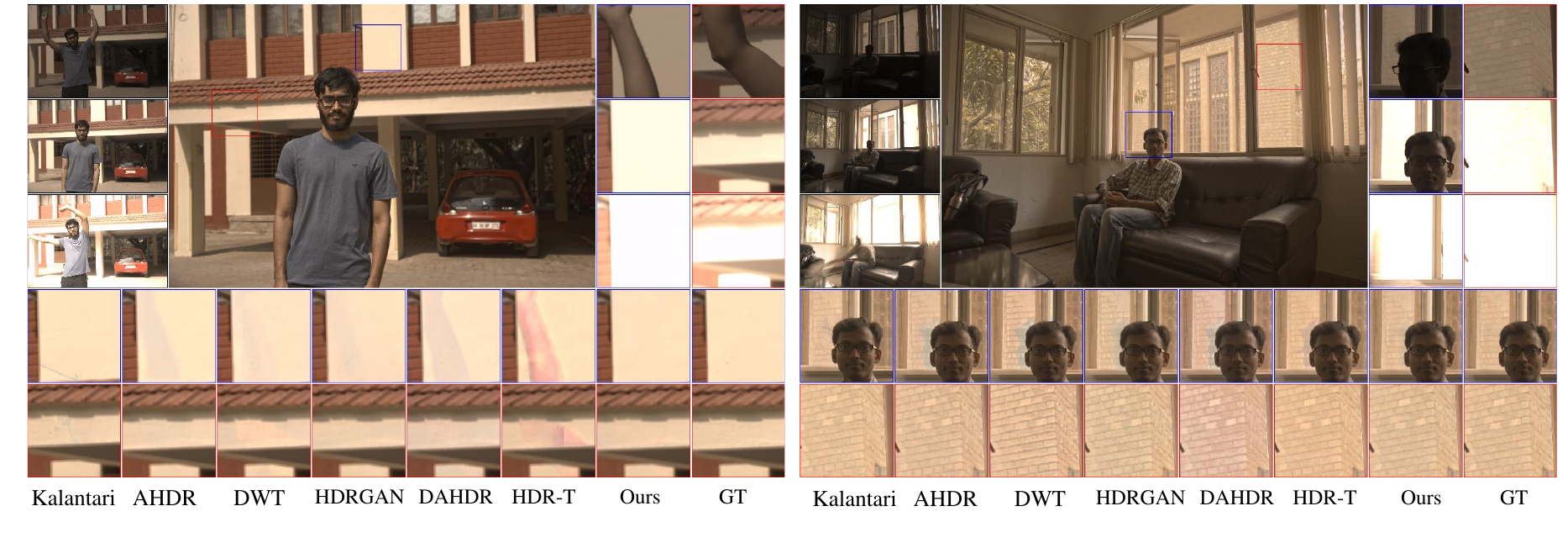}
	\vspace{-6mm}
	\caption{Comparison of qualitative results of the proposed method with the state-of-the-art methods on Prabhakar dataset.  Zoom in for better observation.}
	\label{gen_vision}
	\vspace{-7mm}
\end{figure*}

\section{Experiments and Analysis}
In the following, we first introduce the datasets and implementation details. Then, we compare the proposed IFT architecture with the previous state-of-the-arts on HDR reconstruction task. In the end, we ablate the important design elements of IFT.
\subsection{Implementation Details}
We test the performance of our IFT on two well-known HDR datasets, including Kalantari \cite{kalantari2017deep} and Prabhakar \cite{prabhakar2019fast} datasets. The Kalantari dataset includes $74$ training samples and $15$ test samples, where each sample contains a ground-truth image and three LDR images with varying exposure values of $\langle-2, 0, +2\rangle$ or $\langle-3, 0, +3\rangle$. The Prabhakar dataset comprises $466$ training samples and $116$ test samples, where each sample has a ground-truth image and three LDR images with varying exposure values of $\langle-3, 0, +3\rangle$ or $\langle-4, 0, +4\rangle$. In the training stage, the training samples are cut into $256 \times 256$ image patches, which are then randomly flipped and rotated for data augmentation.  

In order to evaluate the proposed IFT objectively and comprehensively, we adopt three evaluation indicators from different perspectives, including peak signal-to-noise ratio (PSNR), structural similarity index (SSIM) and HDR-VDP-2. In addition, we not only calculate the PSNR-$\mu$ and SSIM-$\mu$ scores between the tonemapped reconstructed HDR results and the corresponding ground truths, but also compute the PSNR-$l$ and SSIM-$l$ values between the reconstructed results in linear domain and the corresponding ground truths. 

The proposed IFT framework is implemented by PyTorch and the experiments are conducted on four NVIDIA 2080Ti GPUs. We adopt the Adam optimizer with default settings $\beta_1 = 0.9$, $\beta_2 = 0.999$, and $\epsilon=10^{-8}$. The learning rate is initially set to $2\times10^{-4}$, which is exponentially decayed to $10^{-6}$ by power $0.3$. The size of window, embedding dimension, and the number of attention heads are set to $8$, $16$, and $5$, respectively. 
To demonstrate the advantages of the proposed IFT, we compare it with $13$ the-state-of-the art HDR methods, including the shallow methods Sen \cite{sen2012robust}, Hu \cite{hu2013hdr}, the deep CNN methods Kalantari \cite{kalantari2017deep}, Wu \cite{wu2018deep}, SCHDR \cite{prabhakar2019fast}, AHDR \cite{yan2019attention}, DWT \cite{dai2021waveletbased}, HDRGAN \cite{HDR_GAN_2021}, DAHDR \cite{yan2022dual}, and the transformer-based methods  SwinIR \cite{liang2021swinir}, HDR-Transformer \cite{liu2022ghost} and Song \cite{song2022selective}. 
\begin{table*}[!t] 	
	\centering
	\vspace{-5mm}
	\caption{Quantitative comparison on Kalantari dataset and Probhakar dataset. %
		\textbf{Bold} number is the best performance in all HDR deghosting baselines.} 	\label{tab:tab1} 	\setlength{\tabcolsep}{1.2mm}{
		\begin{small}\resizebox{1.0\textwidth}{!}{ 
				\begin{tabular}{c|ccccc|ccccc} \toprule[1pt]
					\multirow{2}*{Method}  &  \multicolumn{5}{c|}{Kalantari Testset~\cite{kalantari2017deep}} &   \multicolumn{5}{c}{Prabhakar Testset~\cite{prabhakar2019fast}} \\
					\cline{2-11} & PSNR-$\mu$  & SSIM-$\mu$ & PSNR-$L$ & SSIM-$L$ & HDR-VDP-2 & PSNR-$\mu$  & SSIM-$\mu$ & PSNR-$L$ & SSIM-$L$ & HDR-VDP-2 \\
					\midrule[0.8pt] 
					Sen~\cite{sen2012robust} & 40.95 & 0.9085 &	38.31 & 0.9749 & 60.54 &  32.78 & 0.9740 & 30.50  & 0.9749  & 58.41 \\
					
					Hu\cite{hu2013hdr} & 32.19 & 0.9716 & 30.84 & 0.9511 & 57.83 & 30.82 & 0.9710 & 28.87  & 0.9564  & 59.68\\
					
					Kalantari\cite{kalantari2017deep} & 42.74 & 0.9877 & 41.25 & 0.9845 & 64.63 & 35.34 & 0.9782 & 32.08  & \textbf{0.9818}  & 62.91\\
					
					Wu\cite{wu2018deep} & 41.64 & 0.9869 & 40.91 & 0.9847 & 58.37 & 31.31 & 0.9733 & 30.72  & 0.9518  & 62.44\\
					
					SCHDR\cite{prabhakar2019fast} & 40.47 & 0.9931 & 39.68 & 0.9899 & 62.62 & 30.57 & 0.9715 & 31.44  & 0.9722  & 62.44\\
					
					AHDR\cite{yan2019attention} & 43.62 & {0.9956} & 41.03 & 0.9903 & 64.85 & 33.72 & {\color{black} {0.9789}} & 31.83  & 0.9674  & 62.34\\
					
					AHDR$-IFT^*$ & 43.84 & 0.9954 & 41.06 & 0.9903 & 64.88 & 35.98 & {\color{black} {0.9635}} & 31.97  & 0.9575  & 62.54\\
					
					NHDRNet\cite{yan2020deep} & 42.48 & 0.9942 & 40.20 & 0.9889 & 63.16 & 33.09 & 0.9597 & 28.88  & 0.9389  & 59.93\\
					
					DWT\cite{dai2021waveletbased} & 43.91 & \textbf{0.9957} & 41.22 & \textbf{0.9907} & 65.32 & {35.57} & {0.9811} & {33.02}  & {\color{black} {0.9779}}  & {63.47}\\
					
					HDRGAN\cite{HDR_GAN_2021} & 43.92 & 0.9865 & 41.57 & {0.9905} & 65.45 & {\color{black} {35.20}} & \textbf{0.9829} & {\color{black} {30.92}}  & {\color{black} {0.9717}}  & \color{black} {62.95}\\

					DAHDR~\cite{yan2022dual} & 43.84 & {0.9956} & 41.31 & {0.9905} & 64.68 & 35.34 & {\color{black} {0.9798}} & 32.11  & {\color{black} {0.9784}}  & 61.95\\
					
					SwinIR~\cite{liang2021swinir} & 43.42 & 0.9882 & 41.68 & 0.9861 & 64.52 &-&-&-&-&\\
					
					HDR-T\cite{liu2022ghost} & 44.21 & 0.9918 & 42.17 & 0.9889 & 65.60 & {32.24} &  0.9485 & {30.40} & 0.9043 & {62.48}  \\
					
					Song~\cite{song2022selective} & 44.10 & 0.9909 & 41.70 & 0.9872 & 64.68 
					&-&-&-&-&\\
					
					\hline
					{IFT(ours)} & \textbf{44.55} & 0.9914 & \textbf{42.27} & 0.9887 & \textbf{65.63} & \textbf{36.40} & 0.9673 & \textbf{34.05} & 0.9765 & \textbf{66.67} \\		
					
					\bottomrule[1pt] 			
			\end{tabular}} 		
	\end{small} 	} 
\end{table*}

\begin{table}[tb] \small
	\begin{center}
		\caption{Quantitative results of multi-exposure HDR imaging on ablation study.}
		\label{tab:ablation}
		\resizebox{0.7\columnwidth}{!}{
			\begin{tabular}{l | c | c | c | c | c| c}
				\toprule[1pt]
				BL & FGPS & $FGPS_2$ & SCF & PSNR-$\mu$ & PSNR-$L$ & HDR-VDP-2 \\
				\hline
				$\checkmark$ &  &  &  & 43.5637 & 41.8962 & 65.1601
				\\
				$\checkmark$ & $\checkmark$ &  &  & 44.3575 & 42.0472 & 65.4737\\
				$\checkmark$ &  & $\checkmark$ &  & 43.8381 & 41.4522 & 65.2697\\				
				$\checkmark$ &  &  & $\checkmark$ & 44.2364 & 41.7709 & 65.8112\\
				$\checkmark$ & $\checkmark$ &  & $\checkmark$ & 44.5532 & 42.2714 & 65.6296\\
				\bottomrule[1pt]
		\end{tabular}}
	\end{center}
	\vspace{-5mm}
\end{table}
\textbf{Quantitative results.} 
Tables \ref{tab:tab1} provide the quantitative results of different methods on the Kalantari datasets. As we can see, the proposed IFT achieves the best performance in most cases, which demonstrates the effectiveness and superiority of our method. In particular, our IFT obtains the highest PSNR-$\mu$ score of $44.55$ and PSNR-$l$ score of $42.27$ on the Kalantari dataset, which are $0.35$ and $0.11$ higher than the second ones, respectively. Compared with the flow-based methods \cite{kalantari2017deep, prabhakar2019fast} which explicitly align moving objects, the methods \cite{yan2019multi, yan2019attention, yan2020deep} adopting implicit feature alignment always obtain better performance.     In addition, we observe that transformer-based methods usually outperform deep CNN-based approaches, demonstrating the significance and necessity of the attention module to model long-range dependencies for HDR imaging task. It is worth noting that AHDR$_IFT$ replacing AHDR's fusion module with proposed FGPS and SCF module, obtaines a 0.22db improvement in PSNR-$\mu$ compared with AHDR, which demonstrates the effectiveness and superiority of our method. 

\textbf{Qualitative evaluation.}
Figure \ref{main_vision} exhibits the HDR imaging results of existing mainstream methods. The first row shows the input LDR images, our tonemapped HDR result, and the corresponding zoomed LDR patches from top to bottom.  Due to the large object motion in LDR images, artifacts can easily appear when fusing the three LDRs. The second and third row list the compared HDR results, where the two comparison locations are highlighted in blue and red, respectively. Compared with transformer-based model, cnn-based models  have lower ability to suppress the generation of artifacts. In the two scenes shown in Figure \ref{main_vision}, the first five results of cnn-based methods have obvious artifacts. The main potential reason is that the model considers the background area as an overexposed area and captures information from the corresponding positions of other frames for fusion due to the lack of sufficient semantic information. In addition, we observe that the flow-based method (See the blue block in Figure \ref{main_vision} (a) and red block in Figure \ref{main_vision} (b)) does not have hand like artifacts benefiting from the aligned input frames using optical flow before the further merging operation. But inaccurate optical flow estimation leads to incomprehensible artifacts such as red block in Figure \ref{main_vision} (a) and blue block 
in Figure \ref{main_vision} (b). Compared with HDR-T, ours rarely produce the remaining of artifacts and consistent with ground truth in color (See the red block in Figure \ref{main_vision} (g)) due to the global information fusion capability.

\textbf{Evaluation of generalization ability.}
To verify the generalization ability of our network and others, we select test images from  prabhakar datasets to evaluate. Table \ref{tab:tab1} provide the quantitative results of different methods. As can be seen from the table, our method is far superior to other methods and 0.9 db higher in PSNR-$\mu$ and 1.0db in PSNR-$l$ than the second best results. In addition, Figure \ref{gen_vision} shows the visual results of ours and other state-of-the-art methods. From the figure, we can observe that Figure \ref{gen_vision} (b) (c) (d) (e) (f) generate ghosting artifacts in the images with large foreground motion. Although the method based on optical flow does not produce artifacts similar to "hand" and "head",
unpredictable distortions in details appear in the blue block in Figure \ref{gen_vision} (a) and blue block of Figure \ref{gen_vision} (b). In addition, HDR-T shows poor performance as can be see from Figure \ref{gen_vision} (f) and Table \ref{tab:tab1}, which proves its poor robustness.
Our methods, by contrast, produces better results in terms of detail reconstruction and artifact suppression.

\subsection{Ablation Study}
To investigate the effectiveness of each component in our model, we conduct the ablation study on fast global patch searching module (FGPS) , self-cross Fusion network (SCF). As shown in Table \ref{tab:ablation}, we tested the PSNR score of five groups of models with different settings, where BL represents a tiny model only with local reconstruction transformer blocks. 

\begin{figure}[t]
	\centering
	\includegraphics[width=0.9\textwidth]{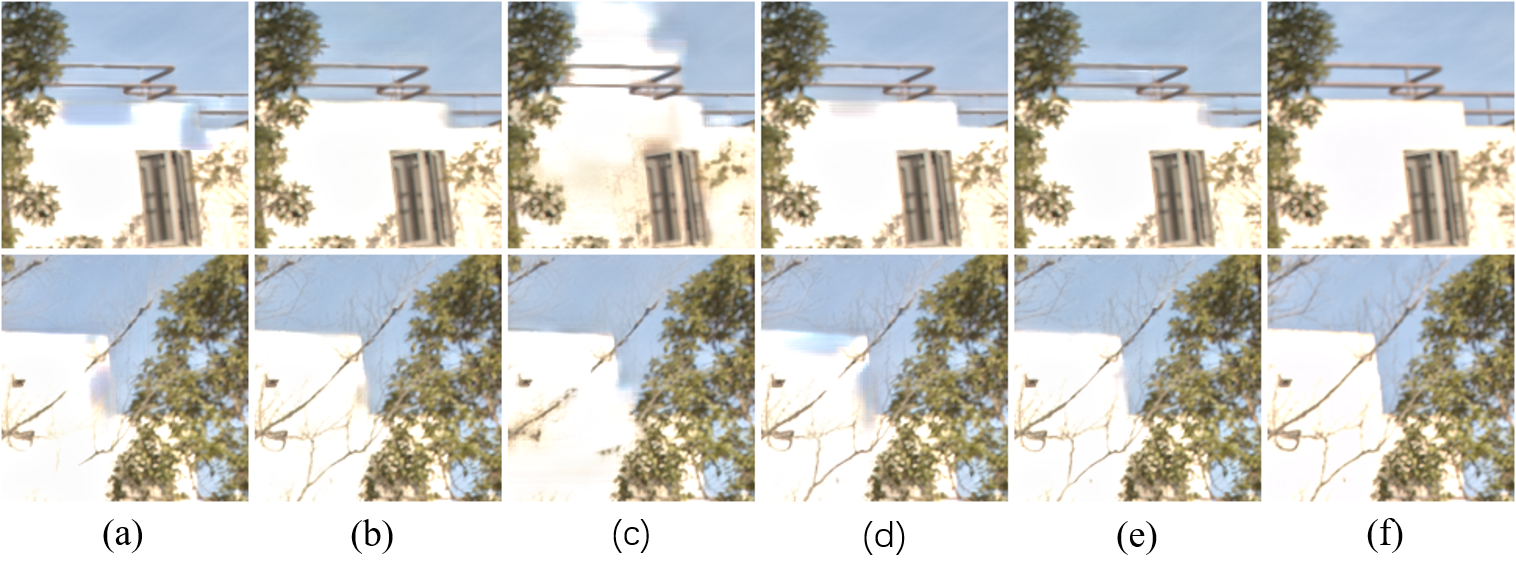}
	\vspace{-3mm}
	\caption{Visual results on ablation study.}
	\label{abso}
	\vspace{-7mm}
\end{figure}

\textbf{Fast global patch searching module.}
This part explores the effect of the proposed fast global patch searching module (FGPS). FGPS aims to search the most similar patches in support frames to the reference frame patches. We add the FGPS to the BL model as a comparison. As shown in Table \ref{tab:ablation}, the FGPS leads to an improvement of 0.71dB in PSNR-$\mu$. The shown in Figure \ref{abso} indicate the importance of the FGPS networks. It is worth noting that in previous experiments, we input the warped features together with the original three features into the local reconstruction module. In addition, we replace the original input features with warped features as input. As shown in Table \ref{tab:ablation}, it turns out that the PSNR-$\mu$ score increases about 0.14dB. Although the warped features are semantically aligned, they are not aligned in the background. Therefore, we finally chose to concat the warped feature and the original feature together as the input of the network.

\textbf{Self-Cross Fusion transformer.}
To see the effect of the proposed self-cross Fusion transformer, we built and trained a BL adding the SCF module. As shown in Table \ref{tab:ablation}, it turns out that the PSNR-$\mu$ scores increases about 0.6db. These results effectively testify the impact of the SCF strategy. The main reason is that the SCF module receives more evidence to choose which spatial information should be focused on for fusion. The results in Figure \ref{abso} (d) visually reflect the importance of the SCF. Finally, the improvement of 1.53dB in PSNR-$\mu$ is achieved by combining FGPS, SCF and BL, which proves the effectiveness of the proposed IFT.

\section{ Conclusion}
In this paper, we have proposed a image fusion transformer for ghost-free imaging, named IFT. Specifically, the proposed IFT contains two modules, one of which responds to search the image patches with the closest semantics to the image patches corresponding to the reference frame in the support frame while the other takes care of multi-frame images fusion. These two parts complement each other. Although the transformer can model global information, it requires a lot of computation. Therefore, using the previous module to find the closest patches and apply the transformer on the size of these patches can greatly reduce the computation. On the other hand, only using the alignment of the semantic patches without the latter module, the incorrect alignment position has a great possibility of producing artifacts and affecting the effect of the model. At last, we carried out detailed experimental comparisons and ablation experiments to verify the effectiveness and rationality of our proposed method, and demonstrated that our method has strong robustness to various scenes. 

In addition to multi-frame HDR reconstruction task, many tasks can also be formulated as the fusion of multi-frame images. In the future, we would attempt to advance the model to solve various multi-frame fusion tasks and design a more general solution.

{\small
	\bibliographystyle{plain}
	\bibliography{egbib2}
}

\end{document}